\documentclass{article}

\usepackage[final]{gpsmdms_2022}
\usepackage[numbers]{natbib}

\usepackage[utf8]{inputenc} %
\usepackage[T1]{fontenc}    %
\usepackage[colorlinks]{hyperref}       %
\usepackage{url}            %
\usepackage{booktabs}       %
\usepackage{amsfonts}       %
\usepackage{nicefrac}       %
\usepackage{microtype}      %
\usepackage{xcolor}         %

\usepackage{amsmath}
\usepackage{mathrsfs}
\usepackage{amsthm}
\usepackage{amssymb}
\usepackage{array}
\usepackage{graphicx} 
\usepackage{geometry}
\usepackage{subcaption}
\usepackage[nameinlink,capitalize]{cleveref}

\usepackage[capitalize,nameinlink]{cleveref}
\crefname{section}{Sec.}{Secs.}
\crefname{appendix}{App.}{Apps.}
\crefname{algorithm}{Alg.}{Algs.}

\usepackage{tikz,pgfplots}
\usetikzlibrary{patterns,shapes.arrows}
\usetikzlibrary{shapes.geometric}

\newlength{\figurewidth}
\newlength{\figureheight}

\pgfdeclareplotmark{mystar}{
    \node[star,star point ratio=2.25,minimum size=7pt,
          inner sep=0pt,draw=white,solid,fill=black,rounded corners=.5pt, anchor=center] {};
}

\newcommand*{\lmlVIf}{\mathcal{L}_{\text{VI}}}

\newcommand*{\lmlEPf}{\mathcal{L}_{\text{EP}}}

\renewcommand{\mid}{\,|\,}

\renewcommand{\paragraph}[1]{{\bf #1}~~}

\usepackage{xspace}

\newcommand{\cf}{\textit{cf.\@}\xspace}

\newcommand{\KL}[2]{\mathrm{D}_{\mathrm{KL}}\big[#1 \,\big\|\, #2 \big]}

\usepackage{bm}
\newcommand{\mathbold}[1]{\bm{#1}}
\newcommand{\mbf}[1]{\mathbf{#1}}

\newcommand{\vtheta}[0]{\mathbold{\theta}}

\newcommand{\vzeta}[0]{\mathbold{\zeta}}

\newcommand{\vf}{\mbf{f}}

\newcommand{\vy}{\mbf{y}}

\newcommand{\diff}{\,\mathrm{d}}

\newcommand{\epsite}[1]{t_{#1}(f_{#1}; \vzeta_{#1})}
\newcommand{\posterior}{p(\vf \mid \vy; \vtheta)}
\newcommand{\prior}{p(\mathbf{f}; \vtheta)}
\newcommand{\likelihoodfactor}{p(y_i \mid f_i ; \vtheta)}
\newcommand{\likelihoodfull}{p(\vy \mid \vf ;\vtheta)}

\title{Towards Improved Learning in Gaussian Processes: The Best of Two Worlds}

\author{
Rui Li \\
Aalto University \\
\texttt{rui.li@aalto.fi} \\
\And
ST John \\
Aalto University \& FCAI\\
\texttt{ti.john@aalto.fi} \\
\And
Arno Solin \\
Aalto University \\
\texttt{arno.solin@aalto.fi} \\
}

\begin{document}

\maketitle

\begin{abstract}
Gaussian process training decomposes into inference of the (approximate) posterior and learning of the hyperparameters. For non-Gaussian (non-conjugate) likelihoods, two common choices for approximate inference are Expectation Propagation (EP) and Variational Inference (VI), which have complementary strengths and weaknesses. While VI's lower bound to the marginal likelihood is a suitable objective for inferring the approximate posterior, it does not automatically imply it is a good learning objective for hyperparameter optimization. We design a hybrid training procedure where the inference leverages conjugate-computation VI and the learning uses an EP-like marginal likelihood approximation. We empirically demonstrate on binary classification that this provides a good learning objective and generalizes better.\looseness-3

\end{abstract}

\section{Introduction}

Gaussian processes (GPs, \cite{gpbook}) provide a principled way of incorporating prior knowledge over functions and quantifying uncertainty. GPs are widely used in a range of applications such as robotics \cite{gpapp1}, numerics \cite{gpapp2} , geostatistics \cite{gpapp3}, and optimization \cite{gpapp4}.
We focus on the non-conjugate case where the training of GP decomposes into two parts: \textit{inferring} the approximate posterior and \textit{learning} the hyperparameters of the model. Variational Inference (VI, \cite{VI}) and Expectation Propagation (EP, \cite{EP}) are two commonly used approximate inference methods for non-conjugate GP models, which have complementary advantages: VI optimizes a lower bound of the marginal likelihood, and is thus easy to implement, straightforward to use, and guaranteed to converge, but is known to underestimate variance \cite{powerep, blei}. EP on the other hand requires implementation-wise tuning per likelihood, and thus is not guaranteed to convergence \cite{epasawayoflife}, but provides a good approximation for the marginal likelihood \cite{05classification, 08classification}. \looseness-2

\begin{figure}
\centering\footnotesize

\setlength{\figurewidth}{0.19\textwidth}
\setlength{\figureheight}{1.\figurewidth}
\pgfplotsset{grid style={dashed,darkgray168},scale only axis,xlabel near ticks,ylabel near ticks, axis on top, tick align=outside, ticklabel style = {font=\tiny}, ytick={-1,1,3,5}, xtick={-1,1,3,5},ylabel style={yshift=-1em, align=center}, grid=both, minor tick num=1}
\begin{tikzpicture}[inner sep=0, outer sep=0]

  \def\data{./picture/tikz/sonar}

  \foreach \x/\name [count=\i] in {MCMC_mean_lml/MCMC,EP_mean_lml/EP,CVI_mean_lml/VI,CVI_mean_lml_ep/Ours} {
    \pgfplotsset{title=\name,ylabel={}}
    \node[anchor=north east] at (1.25*\i*\figurewidth,0) {\input{\data/\x.tex}};
  }

  \node[rotate=90,align=center] at (-.05\figureheight,-.65\figureheight) {\textbf{Marginal likelihood}\\[.6ex]$\log \sigma$};
  \node[rotate=90,align=center] at (-.05\figureheight,-1.85\figureheight) {\textbf{Predictive density}\\[.6ex]$\log \sigma$};

  \foreach \x [count=\i] in {MCMC_nlpd,EP_nlpd} {
    \pgfplotsset{title={},xlabel={$\log \ell$}}
    \node[anchor=north east] at (1.25*\i*\figurewidth,-1.35*\figureheight) {\input{\data/\x.tex}};
  }  

  \pgfplotsset{title={},xlabel={$\log \ell$}}
  \node[anchor=north east] at (1.25*3*\figurewidth,-1.35*\figureheight) {
\begin{tikzpicture}

\definecolor{black20}{RGB}{20,20,20}
\definecolor{black32}{RGB}{32,32,32}
\definecolor{black6}{RGB}{6,6,6}
\definecolor{darkgray154}{RGB}{154,154,154}
\definecolor{darkgray168}{RGB}{168,168,168}
\definecolor{darkorange25512714}{RGB}{255,127,14}
\definecolor{darkslategray47}{RGB}{47,47,47}
\definecolor{darkslategray60}{RGB}{60,60,60}
\definecolor{darkslategray73}{RGB}{73,73,73}
\definecolor{dimgray101}{RGB}{101,101,101}
\definecolor{dimgray113}{RGB}{113,113,113}
\definecolor{dimgray87}{RGB}{87,87,87}
\definecolor{gainsboro222}{RGB}{222,222,222}
\definecolor{ghostwhite249}{RGB}{249,249,249}
\definecolor{gray}{RGB}{128,128,128}
\definecolor{gray141}{RGB}{141,141,141}
\definecolor{lavender235}{RGB}{235,235,235}
\definecolor{lightgray208}{RGB}{208,208,208}
\definecolor{silver181}{RGB}{181,181,181}
\definecolor{silver195}{RGB}{195,195,195}
\definecolor{steelblue31119180}{RGB}{31,119,180}

\begin{axis}[
height=\figureheight,
tick pos=left,
width=\figurewidth,
xmin=-1, xmax=5,
ymin=-1, ymax=5
]
\addplot [draw=none, fill=gray141]
table{%
x  y
1.4 4.38729960387076
1.4107704073557 4.4
1.64919277276499 4.7
1.7 4.76414198622286
1.85456117147986 5
1.7 5
1.4 5
1.1 5
0.8 5
0.788386830715382 5
0.8 4.95378355287469
0.883873708364786 4.7
1.1 4.42479523171893
1.3168440704885 4.4
1.4 4.38729960387076
};
\addplot [draw=none, fill=darkgray154]
table{%
x  y
1.1 3.49915389812451
1.10162268871754 3.5
1.4 3.60603824481895
1.55829926284781 3.8
1.7 3.94062721060176
1.79587282798046 4.1
1.98480378653079 4.4
2 4.42421840669045
2.13682753044301 4.7
2.29683764048468 5
2 5
1.85456117147986 5
1.7 4.76414198622286
1.64919277276499 4.7
1.4107704073557 4.4
1.4 4.38729960387076
1.3168440704885 4.4
1.1 4.42479523171893
0.883873708364786 4.7
0.8 4.95378355287469
0.788386830715382 5
0.5 5
0.419224760450279 5
0.489987555971424 4.7
0.5 4.66224027247765
0.556347039567912 4.4
0.615730467473972 4.1
0.720980945499101 3.8
0.8 3.61554096020367
1.09772570249046 3.5
1.1 3.49915389812451
};
\addplot [draw=none, fill=darkgray168]
table{%
x  y
-0.7 -1
-0.604383781477418 -1
-0.7 -0.746489303349719
-0.720610519336755 -0.7
-0.828131849429229 -0.4
-0.90095170182133 -0.1
-0.952190305863017 0.2
-0.987779701539726 0.5
-1 0.65335967544952
-1 0.5
-1 0.2
-1 -0.1
-1 -0.4
-1 -0.7
-1 -1
-0.7 -1
};
\addplot [draw=none, fill=darkgray168]
table{%
x  y
4.1 -1
4.4 -1
4.7 -1
5 -1
5 -0.7
5 -0.4
5 -0.1
5 -0.0935048891473306
4.99349030554836 -0.1
4.7 -0.393028137094725
4.69299993313807 -0.4
4.4 -0.692088007982958
4.39202965292873 -0.7
4.1 -0.990119637085947
4.08998314052689 -1
4.1 -1
};
\addplot [draw=none, fill=darkgray168]
table{%
x  y
0.8 2.2617682066366
0.91162580737258 2.3
1.1 2.35176549949678
1.31329570923861 2.6
1.4 2.68992511240622
1.53544649710435 2.9
1.7 3.14484654729929
1.73208066207188 3.2
1.90038644195516 3.5
2 3.68103399944567
2.06390413521849 3.8
2.21652439438699 4.1
2.3 4.27547476830926
2.3637020948209 4.4
2.51005256945799 4.7
2.6 4.9079650134735
2.64638869906144 5
2.6 5
2.3 5
2.29683764048468 5
2.13682753044301 4.7
2 4.42421840669045
1.98480378653079 4.4
1.79587282798046 4.1
1.7 3.94062721060176
1.55829926284781 3.8
1.4 3.60603824481895
1.10162268871754 3.5
1.1 3.49915389812451
1.09772570249046 3.5
0.8 3.61554096020367
0.720980945499101 3.8
0.615730467473972 4.1
0.556347039567912 4.4
0.5 4.66224027247765
0.489987555971424 4.7
0.419224760450279 5
0.2 5
0.105551398050781 5
0.109450896201193 4.7
0.118811340980089 4.4
0.147454490326421 4.1
0.2 3.81071668814143
0.202949360241912 3.8
0.25425679957111 3.5
0.295601962602534 3.2
0.370631084108954 2.9
0.494399607207172 2.6
0.5 2.58675726685057
0.760076054899313 2.3
0.8 2.2617682066366
};
\addplot [draw=none, fill=silver181]
table{%
x  y
-0.7 -0.746489303349719
-0.604383781477418 -1
-0.4 -1
-0.1 -1
0.0950179496845851 -1
-0.0954543900588784 -0.7
-0.1 -0.691284302635057
-0.207981979340859 -0.4
-0.304427124525935 -0.1
-0.387394360487999 0.2
-0.4 0.265181382077865
-0.422728446115812 0.5
-0.438707954751541 0.8
-0.448399500355845 1.1
-0.455021841187926 1.4
-0.461143796701067 1.7
-0.469865850481858 2
-0.479270407190252 2.3
-0.479490470261721 2.6
-0.465823808798265 2.9
-0.43905074464233 3.2
-0.400619811299929 3.5
-0.4 3.51007277141998
-0.398549743175615 3.5
-0.317744161959311 3.2
-0.280265043428322 2.9
-0.264924906505515 2.6
-0.253772059507831 2.3
-0.221585783593634 2
-0.123319257536518 1.7
-0.1 1.66009099096269
-0.0479343738814804 1.4
0.0632462997650648 1.1
0.2 0.930486054818022
0.5 0.821296118521705
0.8 0.930333802763269
1.00143650262935 1.1
1.1 1.18158603191376
1.29226980291723 1.4
1.4 1.53664841509238
1.5251054867403 1.7
1.7 1.98824355155727
1.70861036572248 2
1.90635019193901 2.3
2 2.48992727489923
2.0739736884285 2.6
2.23643469283123 2.9
2.3 3.04833423353783
2.39780201037344 3.2
2.54450999825257 3.5
2.6 3.64496335985782
2.69889193565798 3.8
2.8405668279995 4.1
2.9 4.27155636563972
2.9827140449611 4.4
3.12868548340976 4.7
3.2 4.92583621789478
3.24903165302742 5
3.2 5
2.9 5
2.64638869906144 5
2.6 4.9079650134735
2.51005256945799 4.7
2.3637020948209 4.4
2.3 4.27547476830926
2.21652439438699 4.1
2.06390413521849 3.8
2 3.68103399944567
1.90038644195516 3.5
1.73208066207188 3.2
1.7 3.14484654729929
1.53544649710435 2.9
1.4 2.68992511240622
1.31329570923861 2.6
1.1 2.35176549949678
0.91162580737258 2.3
0.8 2.2617682066366
0.760076054899313 2.3
0.5 2.58675726685057
0.494399607207172 2.6
0.370631084108954 2.9
0.295601962602534 3.2
0.25425679957111 3.5
0.202949360241912 3.8
0.2 3.81071668814143
0.147454490326421 4.1
0.118811340980089 4.4
0.109450896201193 4.7
0.105551398050781 5
-0.1 5
-0.4 5
-0.7 5
-1 5
-1 4.7
-1 4.4
-1 4.1
-1 3.8
-1 3.5
-1 3.2
-1 2.9
-1 2.6
-1 2.3
-1 2
-1 1.7
-1 1.4
-1 1.1
-1 0.8
-1 0.65335967544952
-0.987779701539726 0.5
-0.952190305863017 0.2
-0.90095170182133 -0.1
-0.828131849429229 -0.4
-0.720610519336755 -0.7
-0.7 -0.746489303349719
};
\addplot [draw=none, fill=silver181]
table{%
x  y
2.3 -1
2.6 -1
2.9 -1
3.2 -1
3.5 -1
3.8 -1
4.08998314052689 -1
4.1 -0.990119637085947
4.39202965292873 -0.7
4.4 -0.692088007982958
4.69299993313807 -0.4
4.7 -0.393028137094725
4.99349030554836 -0.1
5 -0.0935048891473306
5 0.2
5 0.5
5 0.8
5 1.1
5 1.4
5 1.6629814891862
4.7370005784032 1.4
4.7 1.3630463475152
4.43689483838454 1.1
4.4 1.06319008621707
4.13664013215788 0.8
4.1 0.763512862226346
3.83603096790092 0.5
3.8 0.464241740024363
3.53459349489178 0.2
3.5 0.165880182176157
3.23126987301406 -0.1
3.2 -0.130498924110863
2.92378519666912 -0.4
2.9 -0.422739972797173
2.60739647299332 -0.7
2.6 -0.706827753222807
2.3 -0.972061157162855
2.26723837041595 -1
2.3 -1
};
\addplot [draw=none, fill=silver195]
table{%
x  y
0.2 -1
0.5 -1
0.8 -1
1.1 -1
1.4 -1
1.7 -1
2 -1
2.26723837041595 -1
2.3 -0.972061157162855
2.6 -0.706827753222807
2.60739647299332 -0.7
2.9 -0.422739972797173
2.92378519666912 -0.4
3.2 -0.130498924110863
3.23126987301406 -0.1
3.5 0.165880182176157
3.53459349489178 0.2
3.8 0.464241740024363
3.83603096790092 0.5
4.1 0.763512862226346
4.13664013215788 0.8
4.4 1.06319008621707
4.43689483838454 1.1
4.7 1.3630463475152
4.7370005784032 1.4
5 1.6629814891862
5 1.7
5 2
5 2.3
5 2.6
5 2.9
5 3.2
5 3.5
5 3.8
5 4.1
5 4.4
5 4.7
5 5
4.7 5
4.4 5
4.1 5
3.8 5
3.5 5
3.24903165302742 5
3.2 4.92583621789478
3.12868548340976 4.7
2.9827140449611 4.4
2.9 4.27155636563972
2.8405668279995 4.1
2.69889193565798 3.8
2.6 3.64496335985782
2.54450999825257 3.5
2.39780201037344 3.2
2.3 3.04833423353783
2.23643469283123 2.9
2.0739736884285 2.6
2 2.48992727489923
1.90635019193901 2.3
1.70861036572248 2
1.7 1.98824355155727
1.5251054867403 1.7
1.4 1.53664841509238
1.29226980291723 1.4
1.1 1.18158603191376
1.00143650262935 1.1
0.8 0.930333802763269
0.5 0.821296118521705
0.2 0.930486054818022
0.0632462997650648 1.1
-0.0479343738814804 1.4
-0.1 1.66009099096269
-0.123319257536518 1.7
-0.221585783593634 2
-0.253772059507831 2.3
-0.264924906505515 2.6
-0.280265043428322 2.9
-0.317744161959311 3.2
-0.398549743175615 3.5
-0.4 3.51007277141998
-0.400619811299929 3.5
-0.43905074464233 3.2
-0.465823808798265 2.9
-0.479490470261721 2.6
-0.479270407190252 2.3
-0.469865850481858 2
-0.461143796701067 1.7
-0.455021841187926 1.4
-0.448399500355845 1.1
-0.438707954751541 0.8
-0.422728446115812 0.5
-0.4 0.265181382077865
-0.387394360487999 0.2
-0.304427124525935 -0.1
-0.207981979340859 -0.4
-0.1 -0.691284302635057
-0.0954543900588784 -0.7
0.0950179496845851 -1
0.2 -1
};
\addplot [semithick, steelblue31119180, mark=*, mark size=2.5, mark options={solid,fill=black,draw=white}]
table {%
1.7 0.5
};
\addplot [semithick, darkorange25512714, mark=triangle*, mark size=2.5, mark options={solid,rotate=180,fill=black,draw=white}]
table {%
1.7 0.5
};
\draw (axis cs:1.9,0.5) node[
  scale=0.5,
  anchor=base west,
  text=black,
  rotate=0.0
]{$-0.505$};
\draw (axis cs:1.9,0.5) node[
  scale=0.5,
  anchor=base west,
  text=black,
  rotate=0.0
]{$-0.505$};
\end{axis}

\end{tikzpicture}};
  \node[align=left, text width=\figurewidth] at (4.5*\figureheight,-1.85\figureheight){\scriptsize VI and Ours have the same predictive density surface because they use the same approximate posterior.};

\end{tikzpicture}
\caption{Log marginal likelihood and predictive density surfaces (normalized by $n$) for the {\sc Sonar} data set by varying kernel magnitude $\sigma$ and lengthscale $\ell$. The colour scale is the same in all plots: $-1.2$~\includegraphics[width=1cm,height=.65em]{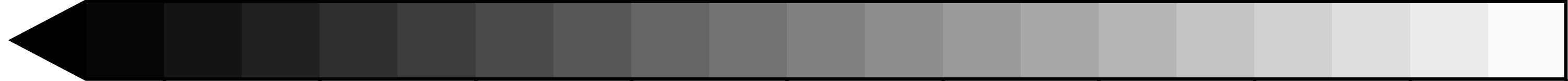}~$-0.2$. Black markers show optimal hyperparameter locations. EP and EP-like marginal likelihood estimation (Ours) match the MCMC baseline better than VI and result in better prediction.}%
\label{fig:sonar_contour}
\tikz[remember picture,overlay]{\draw[blue,shorten >=.2cm,shorten <=.2cm,->] (sepcviu) to [out=200, in = 90] (sepcvil);}
\tikz[remember picture,overlay]{\draw[blue,shorten >=.2cm,shorten <=.2cm,->] (scviu) to [out=200, in=180] (scvil);}
\vspace*{-1.5em}
\end{figure}

When it comes to model performance and generalization to unseen test data, the learning of hyperparameters plays a crucial role. However, it is not entirely clear what is a good learning objective for hyperparameter optimization. This paper fuses the complementary advantages of EP and VI in inference and learning in non-conjugate GP models. We build on \cite{cvi} that provides a link between VI and EP, where the approximate posterior obtained through VI has exactly the same structure as the approximate posterior of EP, and thus provides a plug-in method for obtaining an EP-like marginal likelihood estimate from the VI approximate posterior. Based on this bridge, we propose a hybrid training procedure of GPs where the advantages of VI and EP are combined. The main contributions of this work are: {\em (i)}~We empirically compare the quality of approximated marginal likelihood in EP and VI, and show EP provides better approximation compared with VI (\cref{fig:sonar_contour}); {\em (ii)}~We evaluate the new training procedure in binary classification and demonstrate that it will result in better generalization.

\section{Background}
\label{sec:background}
Gaussian processes are distributions over functions, $f(\cdot) \sim \mathcal{GP}\big(\mu(\cdot), \kappa(\cdot, \cdot)\big)$, fully specified by the mean function $\mu(\cdot)$ and the covariance function $\kappa(\cdot, \cdot)$. Given a data set $\mathcal{D}=(\mathbf{X}, \mathbf{y})=\{(\mathbf{x}_i, y_i)\}_{i=1}^n$ of input--output pairs, the posterior is $\posterior \propto \likelihoodfull \, \prior$, where $\mathbf{f}$ is the vector of function values evaluated at the inputs, $\likelihoodfull = \prod_{i=1}^n \likelihoodfactor$ is the likelihood, and $\vtheta$ denotes the hyperparameters of GP prior and likelihood. For a Gaussian likelihood, the posterior is available in closed form; when the likelihood is non-conjugate, we need to resort to approximate inference.

{\bf Expectation Propagation} (EP, \cite{EP}) seeks to approximate each likelihood term $\likelihoodfactor$ with a {\em site} function $\epsite{i}$. For inference in GPs, the posterior $\posterior$ is approximated by 
\begin{equation}\label{eq: ep_post}\textstyle
	q(\mathbf{f}; \boldsymbol{\theta}, \boldsymbol{\zeta}) = \frac{1}{Z} \prior \prod_{i=1}^n \epsite{i},
\end{equation}
where $\epsite{i}$ is a Gaussian distribution and $Z^{-1}$ is the normalization term. EP approximately minimizes $\KL{\posterior}{q(\mathbf{f}; \boldsymbol{\theta}, \boldsymbol{\zeta})}$. During inference, EP tunes $\boldsymbol{\zeta}_i$ one by one through \looseness-1
\begin{equation}\textstyle\label{eq: ep_infer_obj}
\arg \min_{\boldsymbol{\zeta}_i}	\KL{\likelihoodfactor \prior \prod_{j \neq i} \epsite{j}}{\epsite{i}\prior\prod_{j \neq i} \epsite{j}}.  	
\end{equation}
Learning under the GP paradigm in machine learning typically amounts to finding point estimates for the parameters $\boldsymbol{\theta}$ in the likelihood and kernel by optimizing w.r.t.\ the log marginal likelihood: $\boldsymbol{\theta}^* = \mathrm{arg}\,\max_{\boldsymbol{\theta}} \log p(\mathbf{y}; \boldsymbol{\theta})$. In EP, the log marginal likelihood is directly approximated as \looseness-1
\begin{equation}\label{eq:full_ep_energy}\textstyle
	\log p(\mathbf{y}; \boldsymbol{\theta}) \approx \lmlEPf(\boldsymbol{\zeta}, \boldsymbol{\theta} ) =\log \int \prior \prod_{i=1}^n t_i(f_i; \boldsymbol{\zeta}_i ) \diff\mathbf{f}.
\end{equation}
The direct approximation in \cref{eq:full_ep_energy} is known to lead to a good objective for learning hyperparameters \cite{aki}.
However, the iterative solution of \cref{eq: ep_infer_obj} can be numerically unstable, and is not guaranteed to converge in the general case (see \cite{epasawayoflife} for details).

{\bf Variational Inference} (VI, \cite{VI}) approximates the GP posterior $\posterior$ with a Gaussian distribution $q(\mathbf{f}; \boldsymbol{\xi})=\mathrm{N}(\mathbf{m}, \mathbf{S})$ with variational parameters $\boldsymbol{\xi} = \{\mathbf{m}, \mathbf{S} \}$. VI minimizes the reverse KL $\KL{q(\mathbf{f}; \boldsymbol{\xi} )}{\posterior}$ (\cf the EP section above) by maximizing the following evidence lower bound (ELBO) to the log marginal likelihood:
\begin{equation}\label{eq: full_elbo}\textstyle
\log p(\mathbf{y}; \boldsymbol{\theta}) \geq \lmlVIf(\boldsymbol{\xi}, \boldsymbol{\theta} ) = - \KL{q(\mathbf{f}; \boldsymbol{\xi})}{\prior} + \sum_{i=1}^n \mathbb{E}_{q(f_i; \boldsymbol{\xi}_i )}\big[\log \likelihoodfactor \big] ,
\end{equation}
with respect to variational parameters $\boldsymbol{\xi}$. During inference, as VI optimizes a lower bound of the marginal likelihood, it is guaranteed to converge. In practice, the same lower bound $\lmlVIf(\boldsymbol{\xi}, \boldsymbol{\theta} )$ is used to jointly optimize the variational parameters and a point estimate of the hyperparameters. However, using the lower bound tends to result in biased hyperparameters \cite{05classification, 08classification, powerep}. 

\citet{cvi} introduced a dual parameterization which enables fast natural gradient descent (NGD) on the variational distribution using $\lmlVIf(\boldsymbol{\xi}, \boldsymbol{\theta} )$ as the objective. The Gaussian distribution is part of the exponential family, which means $q(\mathbf{f}) = \mathrm{N}(\mathbf{m}, \mathbf{S}) = \exp \big(\boldsymbol{\eta}^{\top} \mathbf{T}(\mathbf{f})-a(\boldsymbol{\eta})\big)$ where $\boldsymbol{\eta}=[\mathbf{S}^{-1} \mathbf{m},-\frac12 \mathbf{S}^{-1}]$, $\mathbf{T}(\mathbf{f})=[\mathbf{f}, \mathbf{f} \mathbf{f}^{\top}]$, and $\exp(-a(\boldsymbol{\eta}))$ is the normalization term. Instead of using the mean--covariance parameterization $\boldsymbol{\xi} = \{\mathbf{m}, \mathbf{S} \}$ of the Gaussian distribution, we can also use the natural parameters $\boldsymbol{\eta} = \{\mathbf{S}^{-1} \mathbf{m}, -\frac12 \mathbf{S}^{-1} \}$ and expectation parameters ${\boldsymbol{\mu}=\mathbb{E}_{q(\mathbf{f})}[\mathbf{T}(\mathbf{f})]=\{\mathbf{m}, \mathbf{S}+\mathbf{m} \mathbf{m}^{\top}\}}$. 
\citet{cvi} point out that NGD in the natural parameters space has the same computational cost as gradient descent, and the resulting approximate posterior is %
\begin{equation} \label{eq: full_post}\textstyle
  q(\mathbf{f}; \boldsymbol{\lambda},\boldsymbol{\theta}) \propto \prior \prod_{i=1}^n \underbrace{\exp{\left\langle\boldsymbol{\lambda}_i, \mathbf{T}\left(f_i\right)\right\rangle}}_{t_i(f_i; \boldsymbol{\lambda}_i)},
\text { where }
  \boldsymbol{\lambda}_i=\nabla_{\boldsymbol{\mu}_i} \mathbb{E}_{q(f_i; \boldsymbol{\lambda}_i,\boldsymbol{\theta} )}\big[\log \likelihoodfactor \big],
\end{equation}
whose natural parameters $\boldsymbol{\eta}=\boldsymbol{\lambda}_0 + \boldsymbol{\lambda}$ where $\boldsymbol{\lambda}_0=(\boldsymbol{0}, -\frac{1}{2} \kappa (\mathbf{X}, \mathbf{X})^{-1})$ are the natural parameters of the prior $\prior$ and $\boldsymbol{\lambda}$ are the parameters of the likelihood approximation term $t(\mathbf{f})$. Then, we could also parameterize the approximate posterior with $\boldsymbol{\lambda}$, to which we refer as $\boldsymbol{\lambda}$ parameterization. \cref{eq: full_post} has the same form as the EP approximate posterior in \cref{eq: ep_post}. This links EP with VI (see \cite{dual} for details), which is the starting point for our proposed learning objective. \looseness-3

\section{Methods}
VI is the de facto way for approximate inference in GPs, and the de facto method for learning is jointly optimizing the ELBO (\cf\ \cref{eq: full_elbo}) with respect to the variational parameters $\boldsymbol{\xi}$ and the hyperparameters $\boldsymbol{\theta}$. We separate learning from inference by employing a Variational Expectation--Maximization (VEM) procedure similar to \citet{dual}. They alternate between optimizing the variational distribution in the $\boldsymbol{\lambda}$ parameterization and $\boldsymbol{\theta}$ using the following steps at the $t$\textsuperscript{th} iteration:
\begin{equation}
\begin{aligned}
&\text {E-step (inference): } \quad \boldsymbol{\lambda}^{(t+1)} \leftarrow \arg \max _{\boldsymbol{\lambda}} \lmlVIf(\boldsymbol{\lambda}, \boldsymbol{\theta}^{(t)}), \\
&\text {M-step (learning): } \quad \boldsymbol{\theta}^{(t+1)} \leftarrow \arg \max _{\boldsymbol{\theta}} \lmlVIf(\boldsymbol{\lambda}^{(t+1)}, \boldsymbol{\theta}).
\end{aligned}
\end{equation}
The objective for both inference and learning steps is
\begin{equation}\label{eq:viobj}\textstyle
\lmlVIf(\boldsymbol{\lambda}, \boldsymbol{\theta})= -\KL{q(\mathbf{f}; \boldsymbol{\lambda}, \boldsymbol{\theta})}{\posterior}+\sum_{i=1}^n \mathbb{E}_{q(f_i; \boldsymbol{\lambda}_i, \boldsymbol{\theta})}\big[\log \likelihoodfactor \big],
\end{equation}
and the approximate posterior is formed as a product of the prior and Gaussian sites $t_i(f_i; \boldsymbol{\lambda}_i)$ just as in EP (\cref{eq: ep_post}) \cite{cvi}. This allows us to calculate an EP-like estimate of the log marginal likelihood by plugging $\boldsymbol{\lambda}_i$ from \cref{eq: full_post} into $\boldsymbol{\zeta}_i$ in \cref{eq:full_ep_energy}:
\begin{equation}\label{eq:epobj}\textstyle
\lmlEPf(\boldsymbol{\lambda}, \boldsymbol{\theta})= \log \int \prior \prod_{i=1}^n t_i(f_i; \boldsymbol{\lambda}_i) \diff\mathbf{f}.
\end{equation}
As EP tends to provide a better estimation of the marginal likelihood \cite{05classification, 08classification} instead of using \cref{eq:viobj} as the learning objective for hyperparameters, we propose to use \cref{eq:epobj}.

\section{Experiments}
We consider binary classification with a Bernoulli likelihood. We use four common small and mid-sized classification data sets such that there is no need for sparse approximation. The exact details of the data sets can be found in \cref{dataset_appendix}. In all experiments we use the isotropic Mat\'ern-$\nicefrac52$ kernel \cite{gpbook} where the hyperparameters are lengthscale $\ell$ and kernel magnitude $\sigma$. 

\begin{figure}
\centering\footnotesize
\setlength{\figurewidth}{0.19\textwidth}
\setlength{\figureheight}{1.\figurewidth}
\pgfplotsset{grid style={dashed,darkgray168},scale only axis,xlabel near ticks,ylabel near ticks, axis on top, tick align=outside, ticklabel style = {font=\tiny}, ytick={-1,1,3,5}, xtick={-1,1,3,5},ylabel style={yshift=-1em, align=center}, grid=both, minor tick num=1}
\begin{tikzpicture}[inner sep=0, outer sep=0, remember picture]

  \def\data{./picture/tikz/ionosphere}

  \foreach \x/\name [count=\i] in {MCMC_mean_lml/MCMC,EP_mean_lml/EP,CVI_mean_lml/VI,CVI_mean_lml_ep/Ours} {
    \pgfplotsset{title=\name,ylabel={}}
    \node[anchor=north east] at (1.25*\i*\figurewidth,0) {\input{\data/\x.tex}};
  }

  \node[rotate=90,align=center] at (-.05\figureheight,-.65\figureheight) {\textbf{Marginal likelihood}\\[.6ex]$\log \sigma$};
  \node[rotate=90,align=center] at (-.05\figureheight,-1.85\figureheight) {\textbf{Predictive density}\\[.6ex]$\log \sigma$};

  \foreach \x [count=\i] in {MCMC_nlpd,EP_nlpd} {
    \pgfplotsset{title={},xlabel={$\log \ell$}}
    \node[anchor=north east] at (1.25*\i*\figurewidth,-1.35*\figureheight) {\input{\data/\x.tex}};
  }  

  \pgfplotsset{title={},xlabel={$\log \ell$}}
  \node[anchor=north east] at (1.25*3.5*\figurewidth,-1.35*\figureheight) {
\begin{tikzpicture}

\definecolor{black20}{RGB}{20,20,20}
\definecolor{black32}{RGB}{32,32,32}
\definecolor{black6}{RGB}{6,6,6}
\definecolor{darkgray154}{RGB}{154,154,154}
\definecolor{darkgray168}{RGB}{168,168,168}
\definecolor{darkorange25512714}{RGB}{255,127,14}
\definecolor{darkslategray47}{RGB}{47,47,47}
\definecolor{darkslategray60}{RGB}{60,60,60}
\definecolor{darkslategray73}{RGB}{73,73,73}
\definecolor{dimgray101}{RGB}{101,101,101}
\definecolor{dimgray113}{RGB}{113,113,113}
\definecolor{dimgray87}{RGB}{87,87,87}
\definecolor{gainsboro222}{RGB}{222,222,222}
\definecolor{ghostwhite249}{RGB}{249,249,249}
\definecolor{gray}{RGB}{128,128,128}
\definecolor{gray141}{RGB}{141,141,141}
\definecolor{lavender235}{RGB}{235,235,235}
\definecolor{lightgray208}{RGB}{208,208,208}
\definecolor{silver181}{RGB}{181,181,181}
\definecolor{silver195}{RGB}{195,195,195}
\definecolor{steelblue31119180}{RGB}{31,119,180}

\begin{axis}[
height=\figureheight,
tick pos=left,
width=\figurewidth,
xmin=-1, xmax=5,
ymin=-1, ymax=5
]
\addplot [draw=none, fill=gray141]
table{%
x  y
1.4 4.38729960387076
1.4107704073557 4.4
1.64919277276499 4.7
1.7 4.76414198622286
1.85456117147986 5
1.7 5
1.4 5
1.1 5
0.8 5
0.788386830715382 5
0.8 4.95378355287469
0.883873708364786 4.7
1.1 4.42479523171893
1.3168440704885 4.4
1.4 4.38729960387076
};
\addplot [draw=none, fill=darkgray154]
table{%
x  y
1.1 3.49915389812451
1.10162268871754 3.5
1.4 3.60603824481895
1.55829926284781 3.8
1.7 3.94062721060176
1.79587282798046 4.1
1.98480378653079 4.4
2 4.42421840669045
2.13682753044301 4.7
2.29683764048468 5
2 5
1.85456117147986 5
1.7 4.76414198622286
1.64919277276499 4.7
1.4107704073557 4.4
1.4 4.38729960387076
1.3168440704885 4.4
1.1 4.42479523171893
0.883873708364786 4.7
0.8 4.95378355287469
0.788386830715382 5
0.5 5
0.419224760450279 5
0.489987555971424 4.7
0.5 4.66224027247765
0.556347039567912 4.4
0.615730467473972 4.1
0.720980945499101 3.8
0.8 3.61554096020367
1.09772570249046 3.5
1.1 3.49915389812451
};
\addplot [draw=none, fill=darkgray168]
table{%
x  y
-0.7 -1
-0.604383781477418 -1
-0.7 -0.746489303349719
-0.720610519336755 -0.7
-0.828131849429229 -0.4
-0.90095170182133 -0.1
-0.952190305863017 0.2
-0.987779701539726 0.5
-1 0.65335967544952
-1 0.5
-1 0.2
-1 -0.1
-1 -0.4
-1 -0.7
-1 -1
-0.7 -1
};
\addplot [draw=none, fill=darkgray168]
table{%
x  y
4.1 -1
4.4 -1
4.7 -1
5 -1
5 -0.7
5 -0.4
5 -0.1
5 -0.0935048891473306
4.99349030554836 -0.1
4.7 -0.393028137094725
4.69299993313807 -0.4
4.4 -0.692088007982958
4.39202965292873 -0.7
4.1 -0.990119637085947
4.08998314052689 -1
4.1 -1
};
\addplot [draw=none, fill=darkgray168]
table{%
x  y
0.8 2.2617682066366
0.91162580737258 2.3
1.1 2.35176549949678
1.31329570923861 2.6
1.4 2.68992511240622
1.53544649710435 2.9
1.7 3.14484654729929
1.73208066207188 3.2
1.90038644195516 3.5
2 3.68103399944567
2.06390413521849 3.8
2.21652439438699 4.1
2.3 4.27547476830926
2.3637020948209 4.4
2.51005256945799 4.7
2.6 4.9079650134735
2.64638869906144 5
2.6 5
2.3 5
2.29683764048468 5
2.13682753044301 4.7
2 4.42421840669045
1.98480378653079 4.4
1.79587282798046 4.1
1.7 3.94062721060176
1.55829926284781 3.8
1.4 3.60603824481895
1.10162268871754 3.5
1.1 3.49915389812451
1.09772570249046 3.5
0.8 3.61554096020367
0.720980945499101 3.8
0.615730467473972 4.1
0.556347039567912 4.4
0.5 4.66224027247765
0.489987555971424 4.7
0.419224760450279 5
0.2 5
0.105551398050781 5
0.109450896201193 4.7
0.118811340980089 4.4
0.147454490326421 4.1
0.2 3.81071668814143
0.202949360241912 3.8
0.25425679957111 3.5
0.295601962602534 3.2
0.370631084108954 2.9
0.494399607207172 2.6
0.5 2.58675726685057
0.760076054899313 2.3
0.8 2.2617682066366
};
\addplot [draw=none, fill=silver181]
table{%
x  y
-0.7 -0.746489303349719
-0.604383781477418 -1
-0.4 -1
-0.1 -1
0.0950179496845851 -1
-0.0954543900588784 -0.7
-0.1 -0.691284302635057
-0.207981979340859 -0.4
-0.304427124525935 -0.1
-0.387394360487999 0.2
-0.4 0.265181382077865
-0.422728446115812 0.5
-0.438707954751541 0.8
-0.448399500355845 1.1
-0.455021841187926 1.4
-0.461143796701067 1.7
-0.469865850481858 2
-0.479270407190252 2.3
-0.479490470261721 2.6
-0.465823808798265 2.9
-0.43905074464233 3.2
-0.400619811299929 3.5
-0.4 3.51007277141998
-0.398549743175615 3.5
-0.317744161959311 3.2
-0.280265043428322 2.9
-0.264924906505515 2.6
-0.253772059507831 2.3
-0.221585783593634 2
-0.123319257536518 1.7
-0.1 1.66009099096269
-0.0479343738814804 1.4
0.0632462997650648 1.1
0.2 0.930486054818022
0.5 0.821296118521705
0.8 0.930333802763269
1.00143650262935 1.1
1.1 1.18158603191376
1.29226980291723 1.4
1.4 1.53664841509238
1.5251054867403 1.7
1.7 1.98824355155727
1.70861036572248 2
1.90635019193901 2.3
2 2.48992727489923
2.0739736884285 2.6
2.23643469283123 2.9
2.3 3.04833423353783
2.39780201037344 3.2
2.54450999825257 3.5
2.6 3.64496335985782
2.69889193565798 3.8
2.8405668279995 4.1
2.9 4.27155636563972
2.9827140449611 4.4
3.12868548340976 4.7
3.2 4.92583621789478
3.24903165302742 5
3.2 5
2.9 5
2.64638869906144 5
2.6 4.9079650134735
2.51005256945799 4.7
2.3637020948209 4.4
2.3 4.27547476830926
2.21652439438699 4.1
2.06390413521849 3.8
2 3.68103399944567
1.90038644195516 3.5
1.73208066207188 3.2
1.7 3.14484654729929
1.53544649710435 2.9
1.4 2.68992511240622
1.31329570923861 2.6
1.1 2.35176549949678
0.91162580737258 2.3
0.8 2.2617682066366
0.760076054899313 2.3
0.5 2.58675726685057
0.494399607207172 2.6
0.370631084108954 2.9
0.295601962602534 3.2
0.25425679957111 3.5
0.202949360241912 3.8
0.2 3.81071668814143
0.147454490326421 4.1
0.118811340980089 4.4
0.109450896201193 4.7
0.105551398050781 5
-0.1 5
-0.4 5
-0.7 5
-1 5
-1 4.7
-1 4.4
-1 4.1
-1 3.8
-1 3.5
-1 3.2
-1 2.9
-1 2.6
-1 2.3
-1 2
-1 1.7
-1 1.4
-1 1.1
-1 0.8
-1 0.65335967544952
-0.987779701539726 0.5
-0.952190305863017 0.2
-0.90095170182133 -0.1
-0.828131849429229 -0.4
-0.720610519336755 -0.7
-0.7 -0.746489303349719
};
\addplot [draw=none, fill=silver181]
table{%
x  y
2.3 -1
2.6 -1
2.9 -1
3.2 -1
3.5 -1
3.8 -1
4.08998314052689 -1
4.1 -0.990119637085947
4.39202965292873 -0.7
4.4 -0.692088007982958
4.69299993313807 -0.4
4.7 -0.393028137094725
4.99349030554836 -0.1
5 -0.0935048891473306
5 0.2
5 0.5
5 0.8
5 1.1
5 1.4
5 1.6629814891862
4.7370005784032 1.4
4.7 1.3630463475152
4.43689483838454 1.1
4.4 1.06319008621707
4.13664013215788 0.8
4.1 0.763512862226346
3.83603096790092 0.5
3.8 0.464241740024363
3.53459349489178 0.2
3.5 0.165880182176157
3.23126987301406 -0.1
3.2 -0.130498924110863
2.92378519666912 -0.4
2.9 -0.422739972797173
2.60739647299332 -0.7
2.6 -0.706827753222807
2.3 -0.972061157162855
2.26723837041595 -1
2.3 -1
};
\addplot [draw=none, fill=silver195]
table{%
x  y
0.2 -1
0.5 -1
0.8 -1
1.1 -1
1.4 -1
1.7 -1
2 -1
2.26723837041595 -1
2.3 -0.972061157162855
2.6 -0.706827753222807
2.60739647299332 -0.7
2.9 -0.422739972797173
2.92378519666912 -0.4
3.2 -0.130498924110863
3.23126987301406 -0.1
3.5 0.165880182176157
3.53459349489178 0.2
3.8 0.464241740024363
3.83603096790092 0.5
4.1 0.763512862226346
4.13664013215788 0.8
4.4 1.06319008621707
4.43689483838454 1.1
4.7 1.3630463475152
4.7370005784032 1.4
5 1.6629814891862
5 1.7
5 2
5 2.3
5 2.6
5 2.9
5 3.2
5 3.5
5 3.8
5 4.1
5 4.4
5 4.7
5 5
4.7 5
4.4 5
4.1 5
3.8 5
3.5 5
3.24903165302742 5
3.2 4.92583621789478
3.12868548340976 4.7
2.9827140449611 4.4
2.9 4.27155636563972
2.8405668279995 4.1
2.69889193565798 3.8
2.6 3.64496335985782
2.54450999825257 3.5
2.39780201037344 3.2
2.3 3.04833423353783
2.23643469283123 2.9
2.0739736884285 2.6
2 2.48992727489923
1.90635019193901 2.3
1.70861036572248 2
1.7 1.98824355155727
1.5251054867403 1.7
1.4 1.53664841509238
1.29226980291723 1.4
1.1 1.18158603191376
1.00143650262935 1.1
0.8 0.930333802763269
0.5 0.821296118521705
0.2 0.930486054818022
0.0632462997650648 1.1
-0.0479343738814804 1.4
-0.1 1.66009099096269
-0.123319257536518 1.7
-0.221585783593634 2
-0.253772059507831 2.3
-0.264924906505515 2.6
-0.280265043428322 2.9
-0.317744161959311 3.2
-0.398549743175615 3.5
-0.4 3.51007277141998
-0.400619811299929 3.5
-0.43905074464233 3.2
-0.465823808798265 2.9
-0.479490470261721 2.6
-0.479270407190252 2.3
-0.469865850481858 2
-0.461143796701067 1.7
-0.455021841187926 1.4
-0.448399500355845 1.1
-0.438707954751541 0.8
-0.422728446115812 0.5
-0.4 0.265181382077865
-0.387394360487999 0.2
-0.304427124525935 -0.1
-0.207981979340859 -0.4
-0.1 -0.691284302635057
-0.0954543900588784 -0.7
0.0950179496845851 -1
0.2 -1
};
\addplot [semithick, steelblue31119180, mark=*, mark size=2.5, mark options={solid,fill=black,draw=white}]
table {%
1.7 0.5
};
\addplot [semithick, darkorange25512714, mark=triangle*, mark size=2.5, mark options={solid,rotate=180,fill=black,draw=white}]
table {%
1.7 0.5
};
\draw (axis cs:1.9,0.5) node[
  scale=0.5,
  anchor=base west,
  text=black,
  rotate=0.0
]{$-0.505$};
\draw (axis cs:1.9,0.5) node[
  scale=0.5,
  anchor=base west,
  text=black,
  rotate=0.0
]{$-0.505$};
\end{axis}

\end{tikzpicture}};

\end{tikzpicture}
\caption{Log marginal likelihood surfaces for the {\sc Ionosphere} data set. The colour scale is the same in all plots: $-1.4$~\includegraphics[width=1cm,height=.65em]{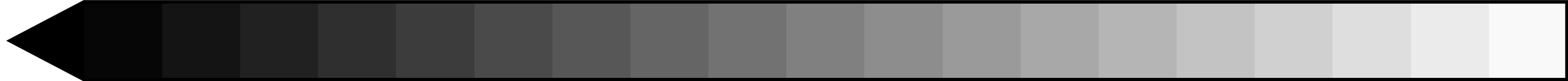}~$-0.1$ (normalized by $n$). Refer to \cref{fig:sonar_contour} for details.}
\label{fig:ionosphere_contour}
\tikz[remember picture,overlay]{\draw[blue, shorten >=.2cm,shorten <=.2cm,->] (epcviu) to[out=250, in = 50] (epcvil);}
\tikz[remember picture,overlay]{\draw[blue, shorten >=.2cm,shorten <=.2cm,->] (cviu) to[out=250, in = 200] (cvil);}
\vspace{-0.5cm}
\end{figure}

\paragraph{Quality of Marginal Likelihood Approximations}
We compare the marginal likelihood approximations of VI, EP and our EP-like VI. As the gold-standard baseline we use Markov Chain Monte Carlo (MCMC). Following \citet{05classification} and \citet{08classification}, we use Annealed Importance Sampling (AIS, \cite{AIS}) to obtain an MCMC estimate of the marginal likelihood. The implementation details can be found in \cref{AIS_detail}.
For each method, we estimate log marginal likelihood on a $21 \times 21$ grid of values for the log hyperparameters $\log \boldsymbol{\theta}=(\log \ell, \log \sigma)$ based on one single fold. We compare different methods by plotting the contour shapes on the hyperparameters grid. As shown in \cref{fig:sonar_contour} on {\sc Sonar} and \cref{fig:ionosphere_contour}  on {\sc Ionosphere} (see \cref{fig:usps_contour,fig:diabetes_contour} in \cref{more_contour} for {\sc USPS} and {\sc Diabetes}), the marginal likelihood estimation of EP closely matches the MCMC baseline, whereas that of VI looks clearly different. Notably, when we estimate the marginal likelihood by plugging the site parameters of VI into the marginal likelihood estimation of EP, the contour shapes become much closer to the MCMC result, which means we have an improved marginal likelihood estimation by using the EP-like marginal likelihood estimation from site parameters of VI. We plot the maximal value of the estimated log marginal likelihood and notice that EP and EP-like VI (Ours) are closer to the MCMC result.

\paragraph{Evaluation on Classification Tasks}
The log marginal likelihood is a surrogate to the generalization ability of the model to unseen data. To explicitly evaluate the generalization ability of our hybrid training procedure, we compare test set accuracy and log predictive density against VI (following the setup of \cite{dual}). Our training procedure uses the EP-like marginal likelihood estimate as learning objective while VI uses ELBO. We perform 5-fold cross-validation and the performance on the test set is given in \cref{full_result} (training details can be found in \cref{training_details}). Our training procedure outperforms VI on {\sc Sonar} and {\sc USPS} and gives very similar performance on {\sc Ionosphere} and  {\sc Diabetes}. This demonstrates that an improved learning objective can give us models that generalize better.

\begin{table}[!h]
\centering\footnotesize
\caption{Test set accuracy and log predictive density on different data sets (mean $\pm$ standard deviation). Higher is better. Results that are statistically significantly different under a paired $t$-test ($p=0.05$) are \textbf{bolded}.}

\begin{tabular}{lcccc}
\toprule
& \multicolumn{2}{c}{Accuracy } & \multicolumn{2}{c}{Log Predictive Density}  \\
&VI & Ours & VI & Ours \\
\midrule 
\sc Ionosphere \cite{dataset} & $0.940 \pm 0.016$ & $0.946 \pm 0.016$ & $-0.179 \pm 0.023$ & $-0.176 \pm 0.023$ \\
\sc Sonar \cite{dataset}      & $0.836 \pm 0.036$ & $\mathbf{0.860 \pm 0.034}$ & $-0.353 \pm 0.013$ & $\mathbf{-0.340 \pm 0.015}$  \\
\sc Diabetes \cite{dataset}   & $0.783 \pm 0.015$ & $0.781 \pm 0.013$ & $-0.473 \pm 0.030$ & $-0.473 \pm 0.030$ \\
\sc USPS \cite{gpbook}        & $0.974 \pm 0.010$ & $0.974 \pm 0.010$ & $-0.080 \pm 0.011$ & $\mathbf{-0.077 \pm 0.011}$ \\
\bottomrule
\end{tabular}
\label{full_result}
\end{table}

\section{Conclusion}
The training of GP models decomposes into inference and learning. In the non-conjugate case EP and VI are two widely used approximate inference methods. In this paper, we study the common choice of the learning objective in VI and empirically demonstrate that EP has a better learning objective. Based on this observation and conjugate-computation VI (CVI, \cite{cvi}) which provides a bridge between VI and EP, we design a hybrid training procedure which has the benefits of VI for inference and the benefits of EP for learning---without any added computational complexity. We evaluate the hybrid training procedure on binary classification tasks and demonstrate that it provides a good learning objective and generalizes better.

\clearpage
\newpage

\appendix

\section{Appendix}

\subsection{Markov Chain Monte Carlo Baseline}
\label{AIS_detail}
As in previous work \cite{05classification,08classification}, we use an MCMC approach as the gold-standard baseline for marginal likelihood estimation. Here we recapitulate their annealed importance sampling approach, which defines a sequence of $t=0,1,\ldots,T$ steps:
	$Z_t= \int \likelihoodfull^{\tau(t)} \,\prior \diff \vf$,
where $\tau(t)=(t/T)^4$ (such that $\tau(0)=0$ and $\tau(T)=1$). The marginal likelihood can be rewritten as
\begin{equation}
	p(\vy;\vtheta)=\frac{Z_T}{Z_0}=\frac{Z_T}{Z_{T-1}} \frac{Z_{T-1}}{Z_{T-2}} \cdots \frac{Z_1}{Z_0},
\end{equation}
where $\frac{Z_t}{Z_{t-1}}$ is approximated by importance sampling using samples from $q_t(\vf ) \propto \likelihoodfull^{\tau(t-1)} \, \prior$:
\begin{equation}
\begin{aligned}
\frac{Z_t}{Z_{t-1}} &= \frac{\int \likelihoodfull^{\tau(t)} \, \prior \diff \vf }{Z_{t-1}}
= \int \frac{p(\mathbf{y} \mid \vf ; \vtheta)^{\tau(t)}}{p(\mathbf{y} \mid \vf ; \vtheta)^{\tau(t-1)}} \frac{p(\mathbf{y} \mid \vf ; \vtheta)^{\tau(t-1)} \, \prior }{Z_{t-1}} \diff \vf  \\
& \approx \frac{1}{S} \sum_{s=1}^S p(\mathbf{y} \mid \vf_t^{(s)}; \vtheta)^{\tau(t)-\tau(t-1)}, \quad \text{where} \quad
 \vf_t^{(s)} \sim \frac{\likelihoodfull^{\tau(t-1)}\,\prior }{Z_{t-1}}.
\end{aligned}
\end{equation}
By using a single sample $S=1$ and a large number of steps $T$, the estimation of log marginal likelihood can be written as 
\begin{equation}
	\log p(\vy;\vtheta) = \sum_{t=1}^{T} \log \frac{Z_t}{Z_{t-1}} \approx \sum_{t=1}^{T} \big(\tau(t)-\tau(t-1)\big)\log p(\mathbf{y} \mid \vf_t; \vtheta).
\end{equation}
Following \cite{05classification}, we set $T=8000$ and combine three estimates of log marginal likelihood by their geometric mean. We use the implementation in GPML toolbox \citep{rasmussen2010gaussian} and use Elliptical Slice Sampling \cite{ess} to sample $\vf_t^{(s)}$.

\subsection{Log Marginal Likelihood on {\sc USPS} and {\sc Diabetes} Data Sets}
\label{more_contour}

\begin{figure}[!h]
\setlength{\figurewidth}{0.19\textwidth}
\setlength{\figureheight}{1.\figurewidth}
\pgfplotsset{grid style={dashed,darkgray168},scale only axis,xlabel near ticks,ylabel near ticks, axis on top, tick align=outside, ticklabel style = {font=\tiny}, ytick={-1,1,3,5}, xtick={-1,1,3,5},ylabel style={yshift=-1em, align=center}, grid=both, minor tick num=1}
\begin{tikzpicture}[inner sep=0, outer sep=0]

  \def\data{./picture/tikz/usps}

  \foreach \x/\name [count=\i] in {MCMC_mean_lml/MCMC,EP_mean_lml/EP,CVI_mean_lml/VI,CVI_mean_lml_ep/Ours} {
    \pgfplotsset{title=\name,ylabel={}}
    \node[anchor=north east] at (1.25*\i*\figurewidth,0) {\input{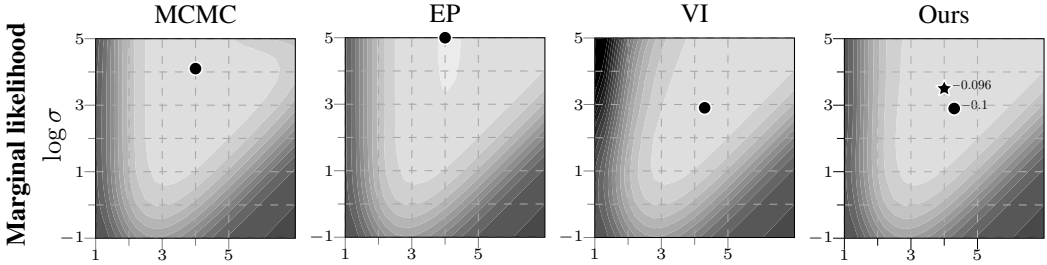}};
  }

  \node[rotate=90,align=center] at (-.05\figureheight,-.65\figureheight) {\textbf{Marginal likelihood}\\[.6ex]$\log \sigma$};
\end{tikzpicture}
\caption{Log marginal likelihood on {\sc USPS} data set. Text annotations are the log predictive density at the corresponding hyperparameter location.}
\label{fig:usps_contour}
\end{figure}

\begin{figure}[!h]
\setlength{\figurewidth}{0.19\textwidth}
\setlength{\figureheight}{1.\figurewidth}
\pgfplotsset{grid style={dashed,darkgray168},scale only axis,xlabel near ticks,ylabel near ticks, axis on top, tick align=outside, ticklabel style = {font=\tiny}, ytick={-1,1,3,5}, xtick={-1,1,3,5},ylabel style={yshift=-1em, align=center}, grid=both, minor tick num=1}
\begin{tikzpicture}[inner sep=0, outer sep=0]

  \def\data{./picture/tikz/diabetes}

  \foreach \x/\name [count=\i] in {MCMC_mean_lml/MCMC,EP_mean_lml/EP,CVI_mean_lml/VI,CVI_mean_lml_ep/Ours} {
    \pgfplotsset{title=\name,ylabel={}}
    \node[anchor=north east] at (1.25*\i*\figurewidth,0) {\input{\data/\x.tex}};
  }

  \node[rotate=90,align=center] at (-.05\figureheight,-.65\figureheight) {\textbf{Marginal likelihood}\\[.6ex]$\log \sigma$};
\end{tikzpicture}
\caption{Log marginal likelihood on {\sc Diabetes} data set. Text annotations are the log predictive density at the corresponding hyperparameter location.}
\label{fig:diabetes_contour}
\end{figure}

\subsection{Data Sets}
\label{dataset_appendix}

The details of data sets we used in experiments are given in \cref{dataset_details}, where $n$ is the number of data points and $d$ is the dimension of each data point.

\begin{table}[h!]
\centering\footnotesize
\caption{Details of data sets.}
\begin{tabular}{lrrl}
\toprule
           & \multicolumn{1}{c}{$n$}    & \multicolumn{1}{c}{$d$}   & Brief description of problem domain \\
\midrule 
\sc Ionosphere \cite{dataset} & 351  & 34  & Classification of radar returns from the ionosphere        \\
\sc Sonar \cite{dataset}     & 208  & 60  & Sonar signals returned by a metal or rock cylinder          \\
\sc Diabetes \cite{dataset}  & 768  & 8   & Predict the outcome on diabetes experiment          \\
\sc USPS \cite{gpbook}   & 1540 & 256 & Binary sub-problem of the USPS handwritten digit data set \\ 
\bottomrule
\end{tabular}
\label{dataset_details}
\end{table}

\subsection{Training Details}
\label{training_details}
We initialize lengthscale $\ell=1$ and magnitude $\sigma=1$. For all data sets, both E-step and M-step are composed of 20 iterations. In the E-step we use natural gradient descent and set the learning rate to $0.1$. In the M-step we use gradient descent and set the learning rate to $0.001$.
\end{document}